\documentclass{bmvc2k}

\title{CSL-YOLO: A New Lightweight Object Detection System for Edge Computing}

\addauthor{Yu-Ming Zhang}{nk108522036@g.ncu.edu.tw}{1}
\addauthor{Chun-Chieh Lee}{jackcclee@cc.ncu.edu.tw}{1}
\addauthor{Jun-Wei Hsieh}{jwhsieh@nctu.edu.tw}{2}
\addauthor{Kuo-Chin Fan}{kcfan@csie.ncu.edu.tw>}{1}

\addinstitution{
 Department of Computer Science and Information Engineering\\
 National Central University\\
 Taoyuan, Taiwan
}
\addinstitution{
 College of Artificial Intelligence and Green Energy\\
 National Chiao Tung University\\
 Hsinchu, Taiwan
}

\runninghead{CSL-YOLO}{A Lightweight Object Detector}

\newcommand{\myeqref}[1]{Eq. (\ref{#1})}
\newcommand{\myfigref}[1]{Fig. \ref{#1}}
\newcommand{\mytabref}[1]{Table \ref{#1}}

\begin{document}

\maketitle

\begin{abstract}
The development of lightweight object detectors is essential due to the limited computation resources. To reduce the computation cost, how to generate redundant features plays a significant role. This paper proposes a new lightweight Convolution method Cross-Stage Lightweight (CSL) Module, to generate redundant features from cheap operations. In the intermediate expansion stage, we replaced Pointwise Convolution with Depthwise Convolution to produce candidate features. The proposed CSL-Module can reduce the computation cost significantly. Experiments conducted at MS-COCO show that the proposed CSL-Module can approximate the fitting ability of Convolution-3x3. Finally, we use the module to construct a lightweight detector CSL-YOLO, achieving better detection performance with only 43\% FLOPs and 52\% parameters than Tiny-YOLOv4.
\end{abstract}

\section{Introduction}
\ \ \ \ Previous research has shown that using deep CNN models can be outstanding performance. Image classifiers or object detectors usually use VGG\cite{vgg}, ResNet\cite{resnet}, and other high FLOPs models as the backbone. In object detection, although the one-stage model led by YOLO\cite{yolov1} is much faster than the two-stage model led by Faster-RCNN\cite{fasterrcnn}, the backbone DarkNet used by YOLO is still a VGG-like or ResNet-like deep CNN model. YOLO has a computing overhead of at least \textasciitilde20k MFLOPs. For edge computing devices, such high computing resource requirements cannot be carried. In the past, the number of model parameters was regarded as a primary goal of model lightweight. However, people soon discovered that the number of parameters is not positively correlated with the running speed. In recent years, more attention has been paid to the number of FLOPs. Although the model's actual running speed is still affected by the framework and OS, and implementation details, FLOPs are still recognized by the mainstream as the most theoretical metrics. In this paper, we use FLOPs as the guide to design a novel lightweight convolution method CSL-Module. We have theoretically proved that CSL-Module is 5 to 7 times faster than convolution-3x3. In our experiments, we have also shown that CSL-Module is faster and performs better than other lightweight convolution methods. Furthermore, CSL-Module was used to construct two highly efficient components, which were finally combined into a new lightweight object detector CSL-YOLO. Compared with other similar lightweight YOLOs, CSL-YOLO has reached the state-of-the-art level.

The rest of the paper is organized as follows: Section 2 briefly reviews the related techniques for lightweight model design, followed by the proposed CSL-Module, CSL-Bone, CSL-FPN, and CSL-YOLO in section 3, the tricks of CSL-YOLO in section 4, the experiments and discussion are presented in section 5, and finally, the conclusion in section 6.

\begin{figure*}
  \centering
  \bmvaHangBox{\fbox{\includegraphics[width=10cm]{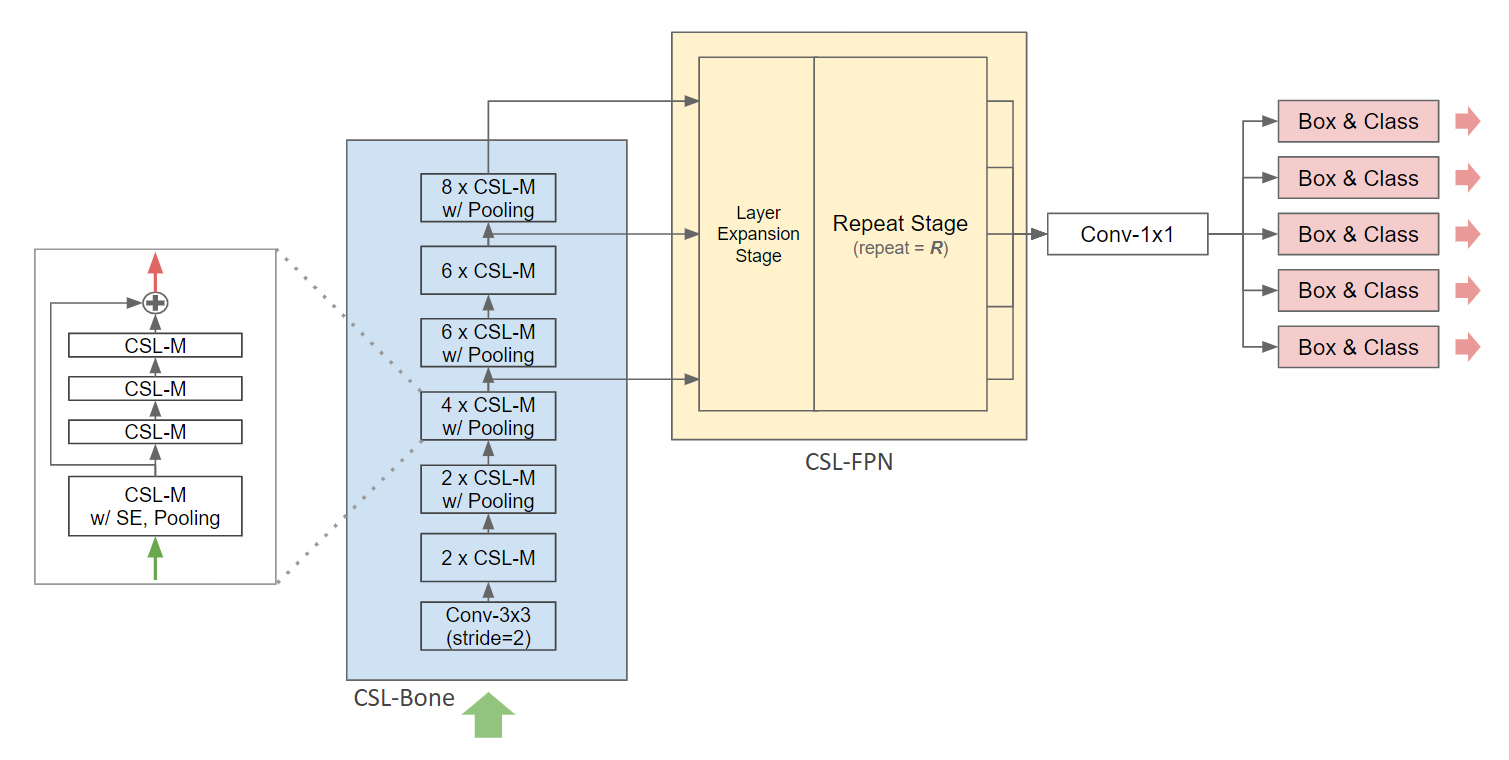}}}\\
  \caption{Overall architecture of CSL-YOLO. the convolution-1x1 is weights-sharing.}
  \label{cslyolo}
\end{figure*}

\section{Related Work}
\ \ \ \ In recent years, many lightweight CNN models have been proposed. In this section, we first review the lightweight convolution methods and backbone, and then we review the lightweight model of the one-stage object detector, especially the SSD\cite{ssd}\cite{retinanet} series and YOLO\cite{yolov1} series.

\subsection{Lightweight Convolution Methods}
\ \ \ \ The previous works have shown that convolution-3x3 is a powerful feature extraction method, but this method is still too expensive for edge computing. Especially in the deep CNN model, many convolution layers are used, which causes unbearable large FLOPs. Depthwise Separable Convolution (DSC) proposed by Google\cite{xception}\cite{mobilenet}. They uses a depthwise convolution to extract feature in space, and then use a pointwise convolution to extract feature in depth. This decoupling convolution method has shown in their experiments that it can approximate the performance of convolution-3x3 with fewer FLOPs. The backbone MobileNet\cite{mobilenet} uses DSC extensively reduces FLOPs but maintains good accuracy. The successor backbone MobileNetv2\cite{mobilenetv2} found that doing nonlinear transformation when the feature dimension is small, it will lose too much useful information. This problem is significant in lightweight models where the feature dimension is strictly controlled. Therefore, they use a pointwise convolution to expand the feature dimension to avoid the information loss caused by nonlinear transformation before the feature-maps passes through the DSC. They call this method Inverted Residual Block (IRB).

In addition to the two methods of decoupling in space and depth to reduce calculations, DSC and IRB, there are also some methods to divide the feature-maps into multiple groups to reduce calculations. ShuffleNet\cite{shufflenet} divides the feature-maps into $G$ groups and passes them through DSC. The formula for FLOPs of convolution is shown as \myeqref{convflops}. $H'$ denotes the height of output. $W'$ denotes the width of output. $C$ denotes the channel of input. $N$ denotes the channel of output. $K$ denotes the kernel size of convolution. According to the it, the feature-maps are grouped so that the input channel $C$ is reduced to $C/G$, the output channel $N$ is reduced to $N/G$, and the FLOPs only have the original $1/G$. CSPNet\cite{cspnet} divides the feature-maps into two halves. The first half of which is generated by convolution, and the other half is directly output after concatenating the first half. GhostNet\cite{ghostnet} discusses this problem more systematically. The proposed Ghost Module uses half of the feature-maps to produce half of the output feature-maps with more expensive transformations, and then uses these feature-maps to generate more redundant feature-maps through cheap linear transformations, and finally concatenates the two halves. These lightweight methods performed well in their experiments.

\begin{equation}
FLOPs_{conv}=H'*W'*C*K^2*N
\label{convflops}
\end{equation}

\subsection{Lightweight Object Detection Method}
\ \ \ \ As generally recognized, object detectors can be classified into two-stage and one-stage. The two-stage model has a stage to generate ROI(Region of Interest), so the accuracy is usually higher, but the speed is also slower than the one-stage model. Although the one-stage model represented by YOLO focuses on the characteristics of real-time detection. However, most FLOPs of the one-stage model are still unacceptable for the tight computing resources of edge computing devices. Therefore, we will review the representative lightweight models in the YOLO and SSD series with extremely low FLOPs.

\subsubsection{SSD Series}
\ \ \ \ SSD is an essential branch of the one-stage detectors, and many designs have led to later detectors. The multi-scale prediction they recommended has indirectly affected the proposal and contribute to the popularization of Feature Pyramid Network(FPN)\cite{fpn}, and they effectively integrated the anchor box proposed by Faster-RCNN\cite{fasterrcnn} to improve the performance significantly. We next introduce a few representative lightweight models in the SSD series. MobileNet-SSD\cite{mobilenet} achieves quite good results with the simple combination of lightweight backbone MobileNet + SSD head. MobileNet-SSDLite further modified the SSD head to make the entire model more lightweight, and the accuracy has improved under the same FLOPs. MobileNetv2-SSDLite\cite{mobilenetv2} replaced the backbone with a lighter MobileNetv2\cite{mobilenetv2} and achieved fantastic speed and accuracy. PeleeNet\cite{peleenet} constructed a DenseNet-like\cite{densenet} lightweight backbone, and at the same time, reduced the output scale of the SSD head to reduce the calculation. They also got impressive results in the experiments.

\subsubsection{YOLO Series}
\ \ \ \ As the one-stage model pioneer, YOLO discards the stage that generates the prior ROI and directly predicts the bounding box. This solution dramatically improves the speed of inference and achieves real-time on the GPU. But YOLO is still too large for embedding devices. YOLO has updated four official versions so far. Although there are corresponding tiny versions from YOLOv1 to YOLOv4\cite{yolov1}\cite{yolov2}\cite{yolov3}\cite{yolov4}, the official updated version focuses more on improving accuracy than speed. Therefore, Tiny-YOLOv1 to Tiny-YOLOv4\cite{yolov1}\cite{yolov2}\cite{yolov3}\cite{yolov4} always follow a similar compression strategy. They remove some convolutional layers or remove some multi-scale output layers in FPN. This strategy achieves a good compression ratio, but it also causes a large loss of accuracy. YOLO-LITE\cite{yololite} follows a more aggressive reduction strategy. YOLO-LITE even only has 482 MFLOPs, but it also loses more considerable accuracy.

\begin{figure}
\centering
\begin{tabular}{cc}
\bmvaHangBox{\fbox{\includegraphics[width=5cm]{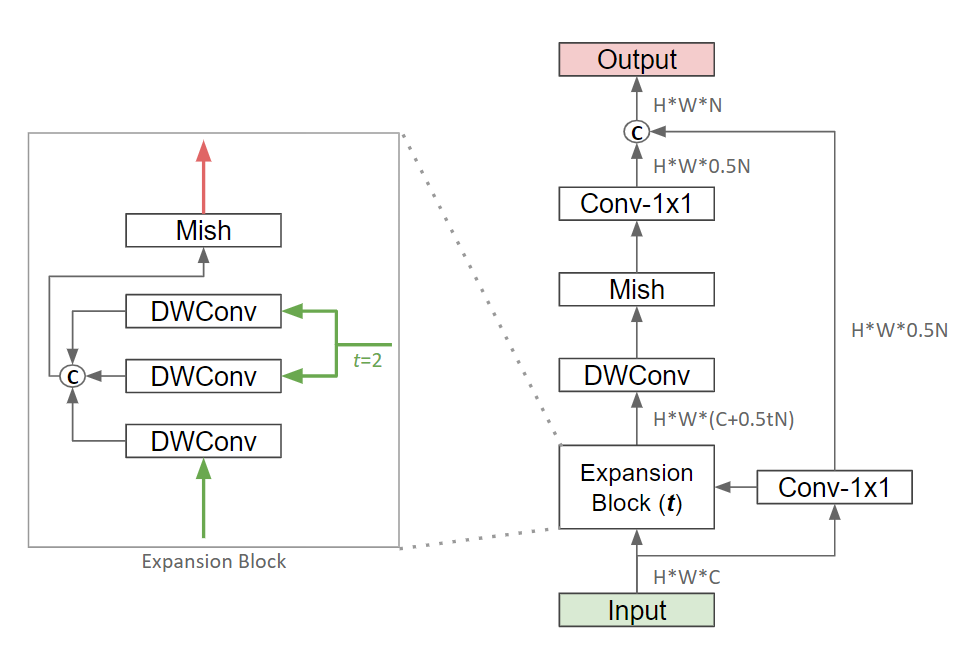}}}&
\bmvaHangBox{\fbox{\includegraphics[width=5cm]{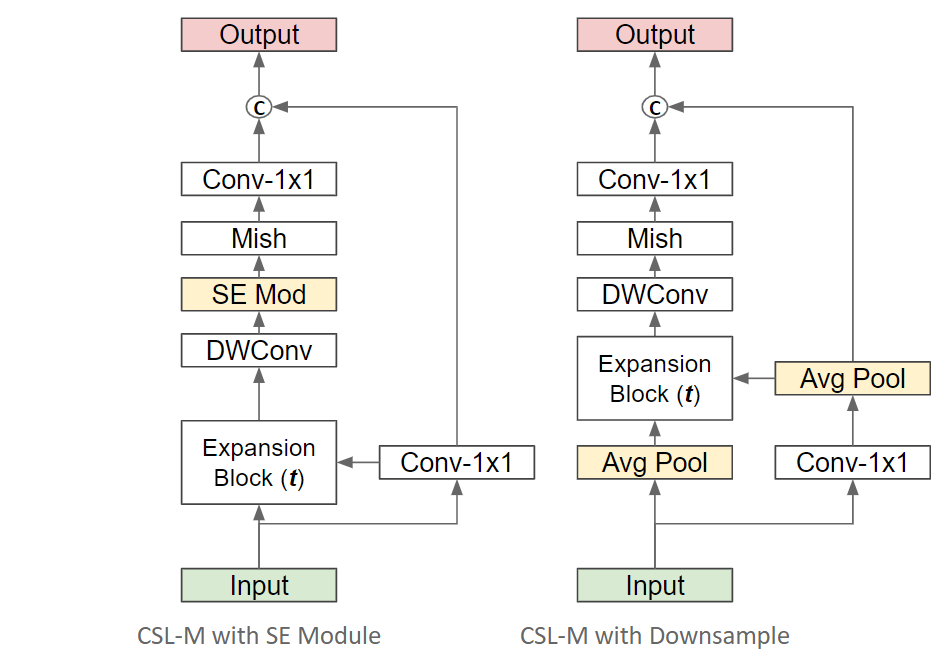}}}\\
(a)&(b)
\end{tabular}
\caption{(a) Overall architecture of CSL-Module. (b) The two variants of CSL-Module are with attention and downsample version.}
\label{cslm}
\end{figure}

\section{Approaches}
\ \ \ \ In this section, we first introduce the Cross-Stage Lightweight (CSL) Module. CSL-Module generates feature-maps with fewer FLOPs, and then we build two lightweight components necessary for object detectors based on CSL-Module.
\subsection{CSL-Module}
\ \ \ \ Previous research has shown that use less computation to generate redundant feature-maps can reduce the FLOPs considerably. CSPNet\cite{cspnet} presents a cross-stage method for solving it, and GhostNet\cite{ghostnet} systematically verifies the effectiveness of the cheap operation in this issue. However, the problem is that the main operation to generate valuable feature-maps is still too expensive for edge computing. We propose dividing the input feature-maps into two branches. The first branch generates the half redundant feature-maps by a cheap operation like GhostNet did; the other branch generates the other half necessary feature-maps by lightweight main operation, then concatenate the two outputs together. The overall architecture is shown in \myfigref{cslm}. The hyperparameter $t$ represents the ratio of feature expanding. We set $t$ as 3 in CSL-Bone, and set $t$ as 2 in else. We insert the SE module\cite{squeezenet} or adaptive average pooling when down-sampling or attention is needed after expansion block. Besides, we use Mish\cite{mish} as the activation function, and in their experiments show that Mish performs better in deep CNN models than ReLU\cite{relu} and Swish\cite{swish}.
\subsubsection{Difference from Existing Methods}
\ \ \ \ The proposed CSL-Module generates half redundant feature-maps by a cheap operation at the skip branch. On the main branch, it is different from CSP Module and Ghost Module. We propose a lightweight main operation to generate the other half necessary feature-maps. In this branch, we design an IRB-like expansion block, using the input feature-maps and output feature-maps of the skip branch to generate intermediate candidate feature-maps by depthwise convolution. One of the great advantages of this block is pointwise convolution-free, and We all know that depthwise convolution has far fewer FLOPs than pointwise convolution. It is different from IRB. IRB uses pointwise convolution to generate candidate feature-maps. The other advantages of this block is it thoroughly considered all currently available features, this can minimizes redundant calculations. Besides, because there is already skip branch, the main branch only needs to generate half of the feature-maps, reducing FLOPs significantly. In general, the proposed CSL-Module reduces FLOPs by cheap operation and cross-stage ideas. On the other hand, we especially lightweight design for the main branch. We replace the convolution layers in VGG-16\cite{vgg} to verify the effectiveness of CSL-Module, and the new models are denoted as IRB-VGG-16, Ghost-VGG-16, and CSL-VGG-16, respectively. We evaluate them on CIFAR-10, the training setting and tricks are all of the same (e.g., flip, affine, mix-up\cite{mixup}, and steps learning rate). From \mytabref{cslmoncifar}, it can be seen that CSL-Module is faster than other advanced lightweight convolution methods, and CSL-Module can more approximate the performance of convolution-3x3. This experiment proves that CSL-Module is a very competitive lightweight convolution method.

\begin{table}
\begin{center}
\begin{tabular}{|c|c|c|}
\hline
& MFLOPs & Acc.(\%)\\
\hline
VGG-16 & 299 & 92.6\\
IRB-VGG-16 & 226 & 92.6\\
Ghost-VGG-16 & 169 & 90.2\\
\textbf{CSL-VGG-16} & \textbf{128} & 92.0\\
\hline
\end{tabular}
\end{center}
\caption{Comparison of CSL-Module with the other lightweight convolution methods on CIFAR-10.}
\label{cslmoncifar}
\end{table}

\subsubsection{Analysis on FLOPs}
\ \ \ \ The analysis is made with the well-known formula \myeqref{convflops}. We assume the input shape and output shape are equal for computational simplicity to compare the speed-up ratio of FLOPs between CSL-Module and convolution-3x3. CSL-Module has a hyperparameter $t$ as same as IRB\cite{mobilenetv2}, representing the expansion ratio of the expansion block. When $t$=2, the speed-up ratio reaches 7.2 times; when $t$=3, it reaches 5.1 times. The whole process is shown in \myeqref{analysissr}
\begin{equation}
\begin{aligned}
FLOPs_{csl}&=(H'*W'*C*0.5N)+t(H'*W'*K^2*0.5N)\\
            &+(H'*W'*C*K^2)\\
            &+(H'*W'*(C+0.5tN)*K^2)\\
            &+(H'*W'*(C+0.5tN)*0.5N)\\
when\ C=&N,\ K=3:\\
&sr=\frac{FLOPs_{conv}}{FLOPs_{csl}}\simeq\frac{9}{1+0.25t}
\label{analysissr}
\end{aligned}
\end{equation}
\subsection{Building Lightweight Components}
\ \ \ \ We propose two lightweight components CSL-Bone and CSL-FPN. These two components are necessary for object detectors. CSL-Bone extracts the features of input image with fewer FLOPs than other backbone models; CSL-FPN predicts the bounding boxes on more different scales efficiently.

\subsubsection{Lightweight Backbone}
\ \ \ \ The proposed CSL-Bone consists of several CSL-Module groups. SE Module\ \cite{squeezenet} has integrated to first CSL-Module in a group to enhance the feature extraction capabilities of the entire group.\ Also, we insert pooling layers for down-sampling at appropriate locations to obtain high-level semantic feature. Finally, CSL-Bone outputs three different scale feature-maps. The overall architecture is shown in \mytabref{cslyolo}. We evaluate CSL-Bone, MobileNetv2, and GhostNet on CIFAR-10 and applied the same training setting too. It can be seen from \mytabref{cslboneoncifar}. Although the CSL-Bone gets lower accuracy than MobileNetv2, but the FLOPs of CSL-Bone just 58.7\% than MobileNetv2. On the other hand, CSL-Bone gets higher accuracy than GhostNet, but only slightly increased FLOPs.

\begin{table}
\begin{center}
\begin{tabular}{|c|c|c|}
\hline
& MFLOPs & Acc.(\%)\\
\hline
MobileNetv2 & 75 & 91.3\\
GhostNet & 40 & 89.3\\
\textbf{CSL-Bone} & 44 & 90.7\\
\hline
\end{tabular}
\end{center}
\caption{Comparison of CSL-Bone with the other lightweight backbons on CIFAR-10.}
\label{cslboneoncifar}
\end{table}

\begin{table}
\begin{center}
\begin{tabular}{|c|c|c|c|}
\hline
& MFLOPs & $AP$ & $AP_{50}$\\
\hline
Vanilla-FPN (conv-3x3) & 416 & 19.0 & 35.8\\
CSL-FPN ($R$=1) & 127 & 18.7 & 35.5\\
CSL-FPN ($R$=2) & 198 & 18.8 & 35.8\\
CSL-FPN ($R$=3) & 268 & 18.8 & \textbf{37.2}\\
CSL-FPN ($R$=4) & 339 & \textbf{19.8} & 37.0\\
CSL-FPN ($R$=5) & 409 & \textbf{19.8} & \textbf{37.2}\\
\hline
\end{tabular}
\end{center}
\caption{The performance of the proposed CSL-FPN with different $R$ on MS-COCO.}
\label{cslfpnonmscoco}
\end{table}

\begin{equation}
\begin{aligned}
&Let\ l\ be\ the\ number\ of\ layers.\\ 
&Let\ k\ be\ the\ number\ of\ anchors\ per\ layer.\\
&Let\ B\ be\ bounding\ boxes\ of\ img.\\
&Let\ A_i\ be\ anchors\ of\ layer_i,\ 0\leq i<l\\
&S=[0,\ \frac{1}{2^{l-1}},\ \frac{1}{2^{l-2}},\ \frac{1}{2^{l-3}},...,\ \frac{1}{2^{l-l}}]\\
&X_i=\{S_i\leq b<S_{i+1}|\forall b \in B\}\\
&C_i=Kmeans(X_i,\ k)\ with\ IoU\ dist\ function.\\
&A_i=\{Center(C_i^0),Center(C_i^1),...,Center(C_i^k)\}
\label{anchorconstraint}
\end{aligned}
\end{equation}

\subsubsection{Lightweight FPN}
\begin{figure}
  \centering
  \bmvaHangBox{\fbox{\includegraphics[width=10cm]{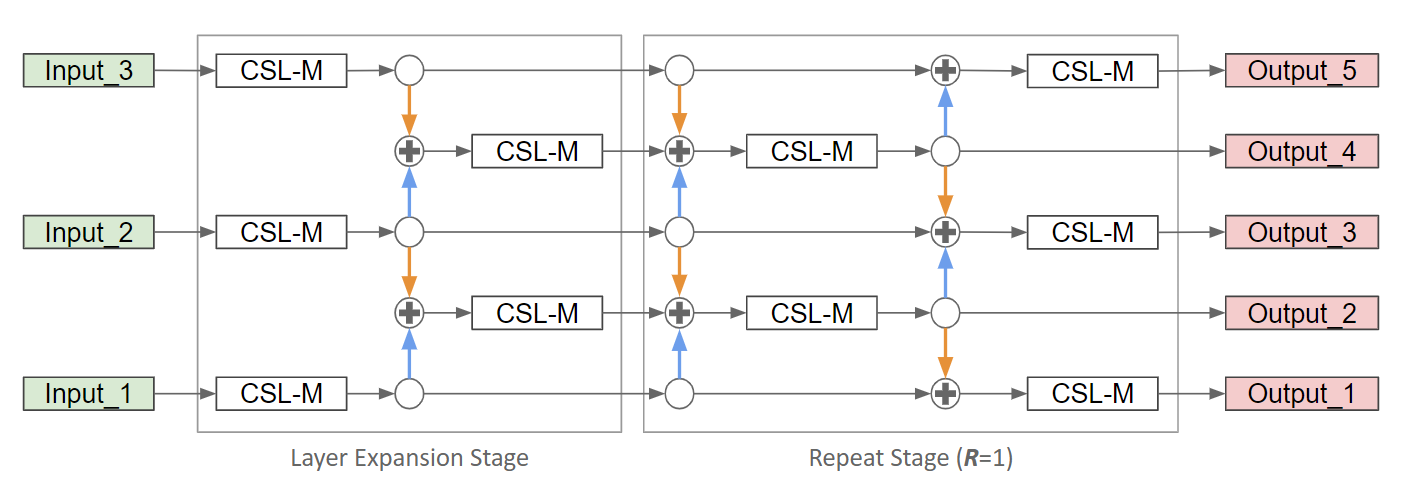}}}\\
  \caption{Overall architecture of CSL-FPN. the layer expansion stage is used to expand the output scale and enhance the continuity between the scales. the repeat stage uses a hyperparameter $R$ to represent the number of times the block is stacked. the blue arrow represents downsampling, the orange arrow represents upsampling, and the white hollow circle represents identity mapping.}
  \label{cslfpn}
\end{figure}

\ \ \ \ Previous studies have shown that large-scale feature-maps have more object details, such as edges, corners, or textures, while small feature-maps have comprehensive semantic understanding. Vanilla FPN\cite{fpn} up-sampling the small feature-maps then fuse them with the large feature-maps. On the other hand, vanilla FPN output three scale feature-maps. It is helpful for the model to detect objects of different sizes. The proposed CSL-FPN first replaces all convolution-3x3 in FPN to CSL-Module. Second, in the expansion stage, a middle-scale layer is formed between two scale layers, and these middle-scale layers can enhance the model's ability to detect objects of different sizes. Third, in the repeat stage,  there are (k)th layer, (k-1)th, and (k+1)th layer for feature fusion at the same time, but each time only odd or even layers are used. For example, there are only the 2nd and 4th layers in the first fusion, and in the second fusion, there are the 1st, 3rd, and 5th layers. In other words, the proposed CSL-FPN has the same number of convolutions as vanilla-FPN but has more feature fusion. The overall architecture is shown in \myfigref{cslfpn}. In our implementation of the proposed CSL-FPN, to make element-wise addition easier, we set the channels of the five output layers to be the same in the layer expansion stage. The repeat stage uses a hyperparameter $R$ to indicate that CSL-FPN has stacked several blocks in total. A larger $R$ can achieve higher AP, but FLOPs also increase. There is a trade-off between speed and performance. We test the best value of $R$ on MS-COCO based on 320x320 CSL-YOLO. \mytabref{cslfpnonmscoco} shows the results. As $R$ gets larger and larger, $AP$ also improved from 18.6\% to 19.8\%, $AP_{50}$ improved from 35.5\% to 37.2\%, but MFLOPs also deteriorated from 127 to 409. After considering the trade-off, we decided to set $R$ to 3.

\section{TRICKS OF CSL-YOLO}
\subsection{Anchors Constraint}
\ \ \ \ The original YOLO series uses K-means with the IoU distance function to cluster the height and width of ground truth, then take the center point as an anchor box. These $k$ anchors generated from $k$ cluster and assign to the output layers of FPN based on their scale. When we expand the output layers from 3 to 5, $k$ also increased from 9 to 15. If we use the above method, then most of these anchors are small-scale due to many small objects in MS-COCO\cite{mscoco}. There will be high-level output layers forced to use small-scale anchors. However, it is well known that high-level feature-maps are not conducive to detect small objects. Therefore, we add the scale limit such as \myeqref{anchorconstraint} before K-means so that the distribution of anchors generated is more in line with each output layer's scale. We conduct experiments in \mytabref{improvementac}, and we can see that the original 3 output layers has worsened after being expanded to 5 output layers. After adding our constraint method, it has improved a considerable AP.

\begin{table}
\begin{center}
\begin{tabular}{|c|c|c|}
\hline
& $AP$ & $AP_{50}$\\
\hline
Output Layers x 3 w/o AC & 17.9 & 36.1\\
Output Layers x 5 w/o AC & 17.2 & 35.1\\
\textbf{Output Layers x 5 w/ AC} & \textbf{18.8} & \textbf{37.2}\\
\hline
\end{tabular}
\end{center}
\caption{Improvement by anchors constraint.}
\label{improvementac}
\end{table}

\subsection{Non-Exponential Prediction}
YOLO series actually predicts the offset of $x$, $y$, $w$, $h$, as follows:

\begin{equation}
\begin{aligned}
&w=w_{anchor}*\mathrm{e}^{w_{pred}},\\
&h=h_{anchor}*\mathrm{e}^{h_{pred}},
\label{logprediction}
\end{aligned}
\end{equation}
where $h_{pred}$ and $w_{pred}$ denotes the offset of the object height and width predicted by the model, $h_{anchor}$ and $w_{anchor}$ denotes the height and width of anchor. Although the log function can limit the prediction of the model range, but the sensitivity of the exponential function makes the width and height quite unstable. Therefore, we removed the log function and let the model directly predict the offset. Then, \myeqref{nonogprediction} can be modified as:
\begin{equation}
\begin{aligned}
&w=w_{anchor}+w_{pred},\\
&h=h_{anchor}+h_{pred}.
\label{nonogprediction}
\end{aligned}
\end{equation}
From \mytabref{nonexppred}, clearly non-exponential prediction can improve 1\textasciitilde2\% AP even under various image sizes. As shown in \myfigref{cslyolo}, all components are then integrated to construct CSL-YOLO.  During inference, the soft-nms\cite{softnms} technique is adopted to penalize overlapping boxes.
\begin{table}
\begin{center}
\begin{tabular}{|c|c|c|c|}
\hline
& Input Size & $AP$ & $AP_{50}$\\
\hline
w/ Exp & 224x224 & 15.0 & 31.1\\
w/ Exp & 320x320 & 18.8 & 37.2\\
w/ Exp & 416x416 & 22.0 & 41.9\\
w/ Exp & 512x512 & 25.1 & 45.7\\
\hline
w/o Exp & 224x224 & \textbf{16.5} & \textbf{32.2}\\
w/o Exp & 320x320 & \textbf{21.4} & \textbf{39.9}\\
w/o Exp & 416x416 & \textbf{24.5} & \textbf{44.0}\\
w/o Exp & 512x512 & \textbf{26.3} & \textbf{46.2}\\
\hline
\end{tabular}
\end{center}
\caption{Improvement by non-exponential prediction.}
\label{nonexppred}
\end{table}

\section{EXPERIMENTS}
\ \ \ \ The proposed CSL-YOLO is not pre-trained on Imagenet\cite{imagenet}. We follow the conclusion of \cite{rethinking}, and train from scratch on MS-COCO. We applied the basic augmentation methods such as flip, rotate, and mix-up\cite{mixup} instead of the fancy methods. We set the batch size as 16, and choosing Adam as the optimizer, The learning rate starts from 1e-3, after 50 epochs, it drops to 1e-4, after 40 epochs, it drops to 1e-5, and finally it is completed after another 20 periods. We train the model on two GTX 1080ti. The entire training process has \textasciitilde800K iterations. Finally, the result of the prediction is shown in \mytabref{cslyoloonmscoco}. Under the input scale of 416x416, the proposed CSL-YOLO uses 3.2M parameters and 1470 MFLOPs to obtain 42.8\% $AP_{50}$, and Tiny-YOLOv4 uses 6.1M parameters and 3450 MFLOPs to obtain 40.2\% $AP_{50}$. It can be said that CSL-YOLO consumes much less time(FLOPs) and space(parameters) than the advanced Tiny-YOLOv4, and can achieve impressive AP performance. Besides, at the input scale of 224x224, compared with the lightest YOLO-LITE, CSL-YOLO still achieves higher AP performance with lower FLOPs.

\begin{table}
\begin{center}
\setlength{\tabcolsep}{0.7mm}{
\begin{tabular}{|c|c|c|c|c|c|c|c|}
\hline
Model & Backbone & Input Size & MFLOPs & Params & $AP$ & $AP_{50}$ &  $AP_{75}$\\
\hline
YOLO-LITE & DarkNet-LITE & 224x224 & 482 & - & - & 12.3 & -\\
Tiny-YOLOv3 & Tiny-DarkNet & 320x320 & 1650 & 8.7M & 14.0 & 29.1 & -\\
Tiny-YOLOv3 & Tiny-DarkNet & 416x416 & 2780 & 8.7M & 16.0 & 33.1 & -\\
Tiny-YOLOv4 & Tiny-CSPDarkNet & 416x416 & 3450 & 6.1M & 21.7 & 40.2 & -\\
\textbf{CSL-YOLO} & CSL-Bone & 224x224 & \textbf{425} & \textbf{3.2M} & \textbf{16.5} & \textbf{32.2} & \textbf{15.0}\\
\textbf{CSL-YOLO} & CSL-Bone & 320x320 & \textbf{869} & \textbf{3.2M} & \textbf{21.4} & \textbf{39.9} & \textbf{20.6}\\
\textbf{CSL-YOLO} & CSL-Bone & 416x416 & \textbf{1470} & \textbf{3.2M} & \textbf{24.5} & \textbf{44.0} & \textbf{24.2}\\
\textbf{CSL-YOLO} & CSL-Bone & 512x512 & \textbf{2223} & \textbf{3.2M} & \textbf{26.3} & \textbf{46.2} & \textbf{26.5}\\
\hline
MobileNet-SSD & MobileNet & 300x300 & 1200 & 6.8M & 19.3 & - & -\\
MobileNet-SSDLite & MobileNet & 320x320 & 1300 & - & 22.2 & - & -\\
MobileNetv2-SSDLite & MobileNetv2 & 320x320 & 800 & 4.3M & 22.1 & - & -\\
PeleeNet & PeleeNet & 304x304 & 1290 & 6.0M & 22.4 & 38.3 & 22.9\\
\hline
\end{tabular}}
\end{center}
\caption{Comparison of CSL-YOLO with the other state-of-the-art lightweight object detection system on MS-COCO.}
\label{cslyoloonmscoco}
\end{table}

\section{CONCLUSION}
\ \ \ \ We proposed a new lightweight convolution method CSL-Module, in the experiment proved that it is competitive compared to other similar methods. We further proposed two components CSL-Bone and CSL-FPN, both of which achieve better performance with fewer FLOPs. We build a new lightweight object detector CSL-YOLO with these components. Besides, we proposed two tricks, a new anchor generation method and a non-exponential prediction to improve AP. Finally, we trained CSL-YOLO from scratch on MS-COCO. From \mytabref{cslyoloonmscoco}, we can see that CSL-YOLO has better AP and fewer FLOPs and parameters than other lightweight YOLOs. In summary, we have three contributions:
\begin{enumerate}
\item We proposed a novel lightweight convolution method CSL-Module that costs fewer FLOPs to approximate the fitting ability of convolution-3x3.
\item We proposed two lightweight components for object detector, CSL-Bone and CSL-FPN which showed excellent results.
\item We effectively integrated these components, and two tricks to improve AP, and finally proposed a state-of-the-art level lightweight detector CSL-YOLO.
\end{enumerate}

\bibliography{egbib}
\end{document}